\documentclass[conference]{IEEEtran}
\IEEEoverridecommandlockouts
\usepackage{cite}
\usepackage{amsmath,amssymb,amsfonts}
\usepackage{algorithmic}
\usepackage{graphicx}
\usepackage{textcomp}
\usepackage{xcolor}
\usepackage{subfigure}
\usepackage{multicol, multirow}
\usepackage{hyperref}
\usepackage{scalerel}
\usepackage{tikz}
\usetikzlibrary{svg.path}
\usepackage{amsmath,amsfonts}
\usepackage{algorithmic}
\usepackage{algorithm}
\usepackage{array}
\usepackage{textcomp}
\usepackage{stfloats}
\usepackage{url}
\usepackage{verbatim}
\usepackage{graphicx}
\usepackage{subfigure}
\usepackage{multicol, multirow}
\usepackage{hyperref}
\usepackage{scalerel}
\usepackage{tikz}
\usetikzlibrary{svg.path}

\newcommand\ddfrac[2]{\frac{\displaystyle #1}{\displaystyle #2}}

\def\BibTeX{{\rm B\kern-.05em{\sc i\kern-.025em b}\kern-.08em
    T\kern-.1667em\lower.7ex\hbox{E}\kern-.125emX}}
\begin{document}

\title{Towards Improved Imbalance Robustness in Continual Multi-Label Learning with Dual Output Spiking Architecture (DOSA)}

\author{\IEEEauthorblockN{1\textsuperscript{st} Sourav Mishra }
\IEEEauthorblockA{\textit{Department of Aerospace Engineering} \\
\textit{Indian Institute of Science}\\
Bangalore, India \\
\texttt{souravmishr1@iisc.ac.in}}
\and
\IEEEauthorblockN{2\textsuperscript{nd} Shirin Dora}
\IEEEauthorblockA{\textit{Department of Computer Science} \\
\textit{Loughborough University}\\
Loughborough, United Kingdom \\
\texttt{s.dora@lboro.ac.uk}}
\and
\IEEEauthorblockN{3\textsuperscript{rd} Suresh Sundaram }
\IEEEauthorblockA{\textit{Department of Aerospace Engineering} \\
\textit{Indian Institute of Science}\\
Bangalore, India \\
\texttt{vssuresh@iisc.ac.in}}
}

\maketitle

\begin{abstract}
Algorithms designed for addressing typical supervised classification problems can only learn from a fixed set of samples and labels, making them unsuitable for the real world, where data arrives as a stream of samples often associated with multiple labels over time. This motivates the study of task-agnostic continual multi-label learning problems. While algorithms using deep learning approaches for continual multi-label learning have been proposed in the recent literature, they tend to be computationally heavy. Although spiking neural networks (SNNs) offer a computationally efficient alternative to artificial neural networks, existing literature has not used SNNs for continual multi-label learning. Also, accurately determining multiple labels with SNNs is still an open research problem. This work proposes a dual output spiking architecture (DOSA) to bridge these research gaps. A novel imbalance-aware loss function is also proposed, improving the multi-label classification performance of the model by making it more robust to data imbalance. A modified F1 score is presented to evaluate the effectiveness of the proposed loss function in handling imbalance. Experiments on several benchmark multi-label datasets show that DOSA trained with the proposed loss function shows improved robustness to data imbalance and obtains better continual multi-label learning performance than CIFDM, a previous state-of-the-art algorithm.
\end{abstract}


\section{Introduction}
Unlike typical scenarios considered in the supervised learning literature, data arrives as a stream of samples in the real world. Continual Learning (CL) algorithms are designed to learn from such data streams \cite{Delange2021}. Generally, CL algorithms are trained on data associated with a single task at a time. Most existing CL algorithms use task identity information for training and inference \cite{16(icarl), 54}. In the real world, a single sample can get associated with several labels over time. For instance, an image shared on social media may receive diverse annotations from various users over time. As a result, a sample may acquire new labels in subsequent tasks while retaining its original labels \cite{dsll}, making it challenging to predict a sample's label set based only on task identity. Further, it is impractical to store all the past samples for replay. Such cases involve more imbalance than their multi-class counterparts \cite{mmd}, making the existing continual learning methods unsuitable for learning from multi-label data streams. Thus, there is a need for models equipped with task-agnostic learning and inference mechanisms, particularly in the context of continual multi-label learning. Ideally, these models should not rely on samples stored from previous tasks for replay. 

The fields of continual learning (CL) and multi-label learning (MLL) have been well-studied in the literature \cite{Delange2021, mllsurvey}. Continual multi-label learning (CMLL) is a more recent research area \cite{dsll, cifdm}. These ANN-based approaches tend to be computationally heavy. At the same time, effectively handling imbalance and learning from samples with no currently available labels continue to be major challenges in CMLL. Although the latter issue has been addressed to some extent in \cite{EBN-MSL}, CMLL performance can be further improved by addressing data imbalance. One way of doing it is to incorporate measures of the model's classification confidence into the loss function \cite{focal-loss}. The loss functions proposed in \cite{EBN-MSL} lack this property.

On the other hand, Spiking Neural Networks (SNNs) have recently emerged as an energy-efficient \cite{energyeffc} alternative to ANNs. Their energy efficiency is attributed to the use of sparse binary events called spikes. Learning algorithms currently available in the literature for training SNNs are inspired by Spike Time Dependent Plasticity (STDP) \cite{stdporiginal, resume, new_stdp} to gradient-based spikeprop \cite{backprop, gdsp, slayer}. Several CL algorithms have also been proposed for SNNs \cite{bayes, local}. However, accurate multi-labeled classification and task-agnostic CMLL with SNNs is still an open research problem.  

To address the above-mentioned research gaps, this work proposes a dual output SNN architecture (DOSA) empowered with an imbalance-aware loss function for improved MLL and task-agnostic CMLL. DOSA also addresses accurate multi-labeled output prediction with SNNs. The proposed loss function ($\mathcal{L}_{fmm}$) incorporates the model's classification confidence to improve robustness towards imbalance. $\mathcal{L}_{fmm}$ trains DOSA in a maximum margin framework while learning a margin value for each class. This trainable margin adapts to the data imbalance, possibly preventing the model from making overly confident predictions. A modified F1 score (inverse-weighted averaged F1 score $F_{iw}$) is developed to evaluate model performance on imbalanced data. The effectiveness of $\mathcal{L}_{fmm}$ in handling data imbalance is first established by MLL experiments on several benchmark multi-label datasets. The Yeast and the Flags datasets yield the maximum improvements on the micro, and the macro averaged F1 scores, respectively. The Yeast dataset also obtains the maximum improvement on the inverse-weighted averaged F1 score. DOSA attains better CMLL performance with $\mathcal{L}_{fmm}$ on the Flags, Virus, Scene, Human, and Eukaryote datasets than CIFDM \cite{cifdm}.

\section{Related Works}
\subsection{Spiking Neural Networks}
SpikeProp \cite{backprop}, one of the earliest supervised learning algorithms for SNNs, uses the difference between the desired and actual spike times to learn the weights through gradient descent. Spike time-dependent plasticity (STDP) \cite{stdporiginal} is a biologically inspired learning rule for SNNs that updates the synaptic weights based only on local information. Improved variants of spikeprop and STDP learning rules have been proposed in \cite{gdsp, resume}. More details on learning algorithms for SNNs are proposed in \cite{spikesurvey}. Most available works in the SNN literature use the leaky integrate and fire neuron model \cite{OMLA, desnn, interpretable}. A recent study proposes a parametric leaky integrate and fire neuron model with a learnable time constant \cite{PLIF}. The proposed architecture uses this neuron model as it performs better than the leaky integrate and fire model. 

Supervised learning literature primarily addresses multi-class classification cases. To predict the $c^{th}$ class in a multi-class scenario with an SNN, a single output neuron can produce $c$ spikes \cite{mst}, or there can be as many output neurons as there are classes to be learned and the $c^{th}$ neuron spikes first \cite{tmmsnn}. However, these approaches cannot be directly used for multi-label classification, which remains an open problem. This paper proposes a dual output SNN architecture (DOSA) for accurate multi-labeled classification to address this issue. 

Continual learning for multi-class problems with SNNs has been studied in \cite{bayes, local}. While \cite{bayes} uses a Bayesian framework to update SNN weights, \cite{local} uses only local updates. However, learning from multi-label data streams with SNNs is still an open research problem. This paper uses DOSA for task-agnostic CMLL.

\subsection{Multi-Label Learning and Continual Learning}
Recent approaches have employed deep neural networks to learn feature and label correlations to improve the multi-label classification performance \cite{137, BP-MLL}. A label correlation-aware feature extraction method for MLL is proposed in \cite{mvmd}. A graph neural network for MLL is proposed in \cite{ML-gcn}. Several other approaches for MLL have been documented in \cite{mllsurvey}. However, these approaches assume the availability of a fixed set of samples and labels. Hence, these may not be suitable for learning from a data stream.

Continual learning from multi-class data has been extensively studied in the literature. The methods are usually categorized into three types - memory-based, regularization-based, and architecture-based \cite{Delange2021}. Memory-based methods avoid catastrophic forgetting by replaying samples stored from previous tasks \cite{45}. Regularization-based methods penalize the updates of different parameters differently based on their importance to the current task \cite{53}. Architecture-based methods grow the model size to fit new tasks or assign task-wise parameters using masks \cite{supermasks}. More details are provided in \cite{Delange2021}. Multi-label datasets usually have more imbalance, and it is possible to find samples with no labels and particular label combinations with no associated samples \cite{mmd, cifdm}. As most memory-based methods rely on sampling the old task data, they might not capture these scenarios applicable to CMLL. Hence, architecture and regularization-based methods must be developed for CMLL. 

Recent studies have emphasized CMLL, showcasing methods like deep streaming label learning (DSLL) \cite{dsll} and continual interactive feature distillation (CIFDM) \cite{cifdm}. CIFDM learns from a stream of samples with continuously emerging labels \cite{cifdm} closely resembling real-world scenarios than DSLL, which maps a fixed set of samples to multiple evolving labels. The CMLL setting presented in CIFDM contains samples with all positive or negative labels, hindering model learning. A bipolar network architecture is proposed in \cite{EBN-MSL} to address this issue. However, the loss functions proposed therein do not distinguish between samples based on the model's ability to classify them. Incorporating classification confidence-based measures into the loss function improves the model's robustness to data imbalance \cite{focal-loss}. Such loss functions are not yet available for bipolar architectures. This work fills this gap by proposing an imbalance-sensitive loss function ($\mathcal{L}_{fmm}$) for improved continual multi-label classification with DOSA.

\section{Method}

\subsection{Continual Multi-Label Learning Problem Formulation}

Let $\mathcal{D} = \{(\mathbf{x}_i, \mathbf{y}_i)\}_{i=1}^N$ be a multi-labeled dataset with input features $\mathbf{x}_i \in \mathcal{R}^m$ and their respective annotations $\mathbf{y} \in \{-1, 1\}^r$. Existing CMLL works divide the datasets into a sequence of tasks $\mathcal{T}_1, \mathcal{T}_2, ...$, with each task presented once for learning \cite{dsll, cifdm}. A task $\mathcal{T}_j$ comprises a collection $\{(\mathbf{x}_k, \mathbf{y}_k)\}_{k=1}^{N_j}$ of $N_j$ samples associated with $\lambda_j$ labels ($\mathbf{y} \in \{-1, 1\}^{\lambda_i}$). The sets of samples and labels in task $\mathcal{T}_j$ are defined as $\mathcal{M}_j$ and $\mathcal{P}_j$, respectively. Consequently, a sample and label may appear in multiple tasks. We assume that a sample may acquire new labels over tasks, but its existing labels remain unchanged. Consequently, the primary objective of CMLL algorithms is to establish a mapping between samples and a dynamically expanding label space, emphasizing learning new labels while minimizing forgetting of previously learned labels. The model architecture, inference, and the proposed focal maximum margin loss function are described in the remainder of this section.

\subsection{Model Architecture and Learning}

\begin{figure*}
    \centering
    \includegraphics[width=0.6\textwidth]{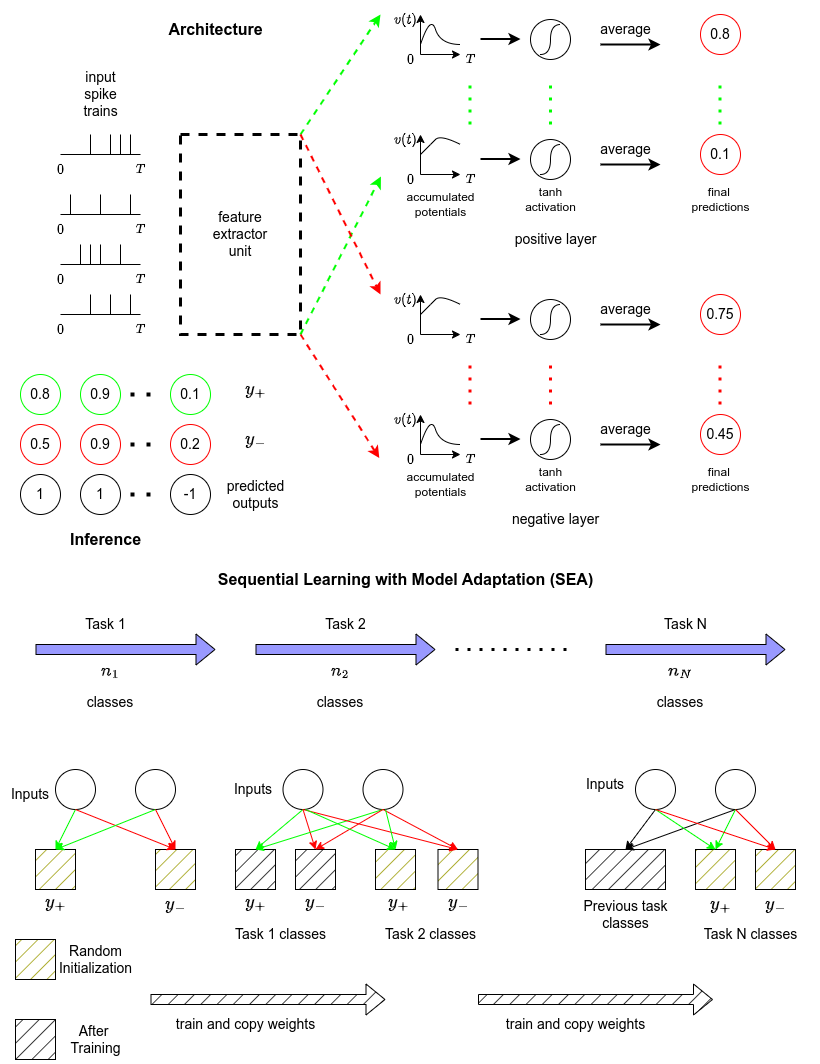}
    \caption{DOSA and inference process. The Sequential Learning with Model Adaptation Setup for Continual Learning with DOSA is shown in the bottom half.}
    \label{fig: arch}
\end{figure*}

Figure \ref{fig: arch} illustrates DOSA. The model has a feature extractor unit comprising multiple fully connected layers and two parallel output layers, positive (green) and negative (red) layers. Hidden layer neurons are modeled using parametric leaky integrate and fire neuron \cite{PLIF}. The output neurons accumulate their potential instead of generating spikes. A tanh activation is applied to these potentials. The resulting values are averaged over the simulation time interval for output prediction. Let $\mathbf{y_k} \in \{-1, 1\}^m$ denote the true labels, $\mathbf{y_{k+}} \in \mathbf{R}^m$ denote the predictions of the positive output layer, $\mathbf{y_{k-}} \in \mathbf{R}^m$ denote those of the negative output layer for the $kth$ sample. As shown in Figure \ref{fig: arch}, inference on multi-labeled data is carried out by comparing $\mathbf{y_{k+}}$ and $\mathbf{y_{k-}}$ in an element-wise manner.

The bottom row of Figure \ref{fig: arch} shows the Sequential Learning with Model Adaptation (SEA) scheme \cite{EBN-MSL} for CMLL with DOSA. The model has no hidden layers in this scheme. The continual learning setting for the datasets is described in section \ref{sec: setups}. We assume that each task is a set of new samples and labels that are disjoint from all previous tasks. Let us assume that the model is trained on all tasks up to $\mathcal{T}_i$. When transitioning to $\mathcal{T}_{i+1}$, the class labels corresponding to $\mathcal{T}_{i+1}$ for the new samples are unavailable. Therefore, we use the model trained on $\mathcal{T}_{i}$ to predict these missing labels for samples in $\mathcal{T}_{i+1}$. A new model is initialized with as many classes as there are in $\mathcal{T}_{i}$ and $\mathcal{T}_{i+1}$. The weights of the old model are copied over to the new model. The weights corresponding only to the classes in $\mathcal{T}_{i+1}$ are randomly initialized. The entire model is then trained on the samples of $\mathcal{T}_{i+1}$, using their augmented labels as ground truths. This process is repeated for all tasks. At any particular task, all the labels seen up to that task are used for training and inference in the SEA mode, making CMLL with DOSA task agnostic.

\subsection{Focal Loss Functions for Maximum Margin Learning}
MLL performance of models can be further improved by handling the imbalance in the data. Incorporating measures of the model's classification confidence in the loss function has been shown to improve its performance on imbalanced data \cite{focal-loss}. However, the loss functions proposed in \cite{EBN-MSL} ($\mathcal{L}_{mm}$) do not distinguish between samples based on the model's classification confidence \ref{eq: eqn 1} and as a result, are not imbalance sensitive. $\mathcal{L}_{mm}$ only seeks to maximize the difference between the scores predicted by the positive and the negative layers. $\mathcal{L}_{mm}$ is minimized at $\mathbf{\zeta_k} = \mathbf{b}$.

\begin{equation}
    \mathcal{L}_{mm}(\mathbf{y_k}, \mathbf{y_{k+}}, \mathbf{y_{k-}}) = \sum_{k = 1}^{N}\exp(||\mathbf{\zeta_k-b}||^2)
    \label{eq: eqn 1}
\end{equation}

Where $\mathbf{\zeta_k} = \mathbf{y_k}\odot(\mathbf{y_{k+}} - \mathbf{y_{k-}})$ and $b$ is the margin. The quantity $\mathbf{\zeta_k}$ defined above can be considered a measure of the model's classification confidence on the $kth$ sample. $\mathcal{L}_{mm}$ considers a fixed margin $b=1$ for each class. $\mathcal{L}_{mm}$ does not really distinguish between samples with high $\mathbf{\zeta_k}$ or low $\mathbf{\zeta_k}$. In order to gain more robustness to imbalance, the model needs to prioritize learning from samples on which it is less confident (low $\mathbf{\zeta_k}$) and vice versa. This mechanism is introduced in the proposed loss function ($\mathcal{L}_{fmm}$) by multiplying $\mathcal{L}_{mm}$ with $\exp(-(\mathbf{\zeta_k-b}))$ as the importance factor. As a result, the model will incur a high loss value on samples where it is less confident (low $\mathbf{\zeta_k}$) and vice versa, bringing imbalance awareness into the learning process.

\begin{equation}
    \mathcal{L}_{fmm}(\mathbf{y_k}, \mathbf{y_{k+}}, \mathbf{y_{k-}}) = \sum_{k = 1}^{N}\exp(-(\mathbf{\zeta_k-b}))\odot(||\mathbf{\zeta_k-b}||^2)
    \label{eq: eqn 2}
\end{equation}


All the classes in a dataset are not imbalanced to the same extent \cite{mmd}. Classes with a higher sample count are encountered more frequently during training, leading to higher confidence predictions for these classes and vice versa. Using the same margin value for all classes might hinder model learning. Hence, $\mathcal{L}_{fmm}$ uses a vector $\mathbf{b}$ of the same length as there are classes in the dataset as the margin. So, there is a unique margin value $b_j$ for each class $j$. Unlike \cite{EBN-MSL} which uses a constant value of $\mathbf{b=1}$ for $\mathcal{L}_{mm}$, $\mathcal{L}_{fmm}$ uses a trainable margin. This influences how strongly the loss on each class is weighed according to their imbalance.

\section{Experimental Settings and Evaluation}
The experimental protocols and the evaluation schemes for multi-label and CMLL and, other relevant implementation details are presented in this section. 

\subsection{Datasets and Continual Multi-Label Learning Settings}
\label{sec: setups}
MLL experiments are performed on ten datasets: yeast \cite{Yeast}, flags \cite{Flags}, birds \cite{Birds}, gpositive, gnegative, plant, human, eukaryote \cite{mvmd}, emotions \cite{Emotions}, and foodtruck \cite{foodtruck}. For these experiments, the datasets are not split into tasks.

CMLL experiments are carried out on five datasets: flags \cite{Flags}, scene \cite{Scene}, virus, human, and eukaryote \cite{mvmd}. Table \ref{tab: Table 1} shows the dataset, task information, and architecture details used in experiments for CMLL. These datasets are divided into tasks following the scheme presented in \cite{cifdm}. During training, each task is a set of new samples and labels distinct from previous ones ($\mathcal{M}_j \cap \mathcal{M}_k = \emptyset$ and $\mathcal{P}_j \cap \mathcal{L}_k = \emptyset$ for $j \neq k$). Therefore, the samples and labels across all tasks add up to the total number of samples and labels in the training set. The test set is divided into tasks for model performance evaluation, keeping all the samples and only distributing the labels across the tasks ($\mathcal{P}_j \cap \mathcal{P}_k = \emptyset$ for $j \neq k$). The labels in the test set are divided between tasks in the same way as in the training set. MLL is a special case of CMLL, with only one task containing the entire dataset.

\begin{table*}[!t]
\centering
\caption{Dataset details, architecture details for MLL, and task-wise division of samples and labels for CMLL following the CIFDM \cite{cifdm} setup.}
\label{tab: Table 1}
\begin{tabular}{ccccccc}
\hline
\textbf{Dataset}   & \textbf{Features} & \textbf{Labels} & \textbf{\begin{tabular}[c]{@{}c@{}}Number of \\ Hidden Layers\end{tabular}} & \textbf{\begin{tabular}[c]{@{}c@{}}Number of \\ Neurons per layer\end{tabular}} & \textbf{\begin{tabular}[c]{@{}c@{}} \# of training samples\\ per task\end{tabular}}                     & \textbf{\# of labels per task}                                                      \\ \hline
\textbf{Plant}     & 440               & 12              & 3                                                                           & 20                                                                              & -                                                                                                 & -                                                                             \\
\textbf{Virus}     & 440               & 6               & -                                                                           & -                                                                               & {[}32, 60, 32{]}                                                                                  & {[}2, 2, 2{]}                                                                 \\
\textbf{Emotions}  & 72                & 6               & 3                                                                           & 10                                                                               & -                                                                                                 & -                                                                             \\
\textbf{Flags}     & 19                & 7               & 2                                                                           & 5                                                                               & {[}43, 43, 43{]}                                                                                  & {[}3, 2, 2{]}                                                                 \\
\textbf{Yeast}     & 103               & 14              & 3                                                                           & 10                                                                              & -                                                                                                 & -                                                                             \\
\textbf{FoodTruck} & 21                & 12              & 2                                                                           & 5                                                                               & -                                                                                                 & -                                                                             \\
\textbf{Scene}     & 294               & 6               & -                                                                           & -                                                                               & {[}405, 403, 403{]}                                                                               & {[}2, 2, 2{]}                                                                 \\
\textbf{Gpositive} & 440               & 4               & 3                                                                           & 20                                                                              & -                                                                                                 & -                                                                             \\
\textbf{Gnegative} & 440               & 8               & 3                                                                           & 20                                                                              & -                                                                                                 & -                                                                             \\
\textbf{Human}     & 440               & 14              & 3                                                                           & 20                                                                              & {[}310, 310, 310, 310, 622{]}                                                                     & {[}2, 2, 2, 2, 3{]}                                                           \\
\textbf{Eukaryote} & 440               & 22              & 3                                                                           & 20                                                                              & \begin{tabular}[c]{@{}c@{}}{[}435, 530, 438, 465, 465, \\ 465, 465, 465, 465, 465{]}\end{tabular} & \begin{tabular}[c]{@{}c@{}}{[}2, 2, 2, 2, 2, \\ 2, 2, 2, 2, 1{]}\end{tabular} \\
\textbf{Birds}     & 260               & 19              & 3                                                                           & 20                                                                              & -                                                                                                 & -                                                                             \\ \hline
\end{tabular}
\end{table*}

\subsection{Performance Measures}

Model performance evaluation in MLL uses the micro, macro, weighted, and inverse-weighted averaged F1 scores as metrics. Let $n_k$ denote the proportion of samples (out of all samples) in class $k$ and $F_k$ denote the F1 score obtained by the model on it. We define the weighted ($F_w$) and inverse weighted averaged F1 ($F_{iw}$) scores on the entire dataset by equations \ref{eq: m1} and \ref{eq: m2}.

\begin{equation}
    F_w = \sum_{k} n_kF_k
    \label{eq: m1}
\end{equation}

\begin{equation}
    F_{iw} = \ddfrac{\sum_{k} F_k/n_k}{\sum_{k}1/n_k}
    \label{eq: m2}
\end{equation}

$F_{iw}$ assigns more importance to imbalanced classes due to inverse weighting with sample proportions $n_k$, serving as an informative metric to evaluate the effectiveness of $\mathcal{L}_{fmm}$. These results are presented in tables \ref{tab: mll_perf} and \ref{tab: imb_perf}. 

Model evaluation on a task during CMLL is carried out in the combined evaluation mode \cite{EBN-MSL} using the macro-averaged F1 score. In this mode, the model is evaluated at the end of training on all tasks $\mathcal{T}_1, \ \mathcal{T}_1, \ ..., \mathcal{T}_{n}$. For evaluating the performance on $\mathcal{T}_i$, all the labels seen up to $\mathcal{T}_i$ (that is, labels in $\mathcal{T}_1,\ \mathcal{T}_2,\ ..., \mathcal{T}_i$) are used \cite{cifdm}. The effect of catastrophic forgetting on model performance is more prominent in the combined evaluation mode, as the model is tested at the end of training on all tasks. Therefore, we use this mode to report the final performance of models.

The combined evaluation mode alone is insufficient for evaluating forgetting and other measures, as provided in \cite{riemann, new_metrics}. Additionally, forgetting in CMLL can be attributed to parameter change and error accumulation by predicting missing labels while transitioning between tasks. Hence, appropriate evaluation modes and performance measures must be developed to accurately quantify the contributions of each of the above effects to overall catastrophic forgetting. Since the paper focuses on improving continual multi-label classification performance by addressing data imbalance, a detailed framework for CMLL evaluation is left for future research.

\subsection{Baselines}
MLL experiments are performed to establish that $\mathcal{L}_{fmm}$ handles imbalance better than $\mathcal{L}_{mm}$. For this reason, DOSA trained with $\mathcal{L}_{mm}$ is used as the baseline. In the CMLL experiments, DOSA trained with $\mathcal{L}_{mm}$ and $b=1$ following the SEA scheme is used as a baseline along with the CIFDM \cite{cifdm} method. As CIFDM \cite{cifdm} outperforms DSLL \cite{dsll} in a more challenging setting, we use CIFDM as a baseline instead of DSLL. We do not compare with \cite{image} as they use a different dataset division scheme and evaluation procedure. However, $\mathcal{L}_{fmm}$ and $\mathcal{L}_{mm}$ can be used with bipolar networks having convolutional feature extractor backbones. As the main focus of this work is developing imbalance-sensitive loss functions for DOSA, we do not use imbalance-aware sampling-based methods such as \cite{ADASYN, SMOTE} as baselines.

\subsection{Hyparameters and Implementation details}
The model has a few hidden layers for MLL experiments (Table \ref{tab: Table 1}). The model has no hidden layers for CMLL as it follows the SEA setup from \cite{EBN-MSL}. This ensures fair comparison with \cite{EBN-MSL} and avoids overfitting, as tasks have fewer samples than the entire dataset. For the same reasons, the embedding dimensions in CIFDM \cite{cifdm} are reduced to 20. All models are trained at a learning rate of 0.001 for 100 epochs. SNNs are implemented using Spikingjelly \cite{PLIF}. The importance factor $\exp(-(\mathbf{\zeta_k-b}))$ is clamped at a minimum value of $0.001$. For all SNNs, the simulation duration $T$ is set to 10 $\text{ms}$ with a time interval of 1 $\text{ms}$. The real-valued input features are converted to spike times using the Poisson-based rate encoding mechanism present in the Spikingjelly library \cite{spikingjelly}. Following \cite{EBN-MSL}, $\mathbf{b}$ is set to $\mathbf{1}$ for $\mathcal{L}_{mm}$. Evaluation metrics are taken from the scikit-learn package. Experiments are carried out on an NVIDIA GeForce RTX 3090 Ti GPU.

\section{Results and Comparative Analysis}
The performance evaluation results of DOSA in MLL and CMLL are presented in this section. MLL experiments are performed first to establish that $\mathcal{L}_{fmm}$ is better at handling imbalance than $\mathcal{L}_{mm}$ before moving on to CMLL. Since DOSA follows the SEA setup for continual learning \cite{EBN-MSL}, it does not have any specific strategies like distillation to address forgetting. Therefore, to first establish that the proposed loss function $\mathcal{L}_{fmm}$ can handle imbalance better than $\mathcal{L}_{mm}$, MLL experiments are performed. Then it is applied to CMLL.

\subsection{Ablation Studies}
\begin{figure}[!t]
    \centering
    \includegraphics[width=0.45\textwidth]{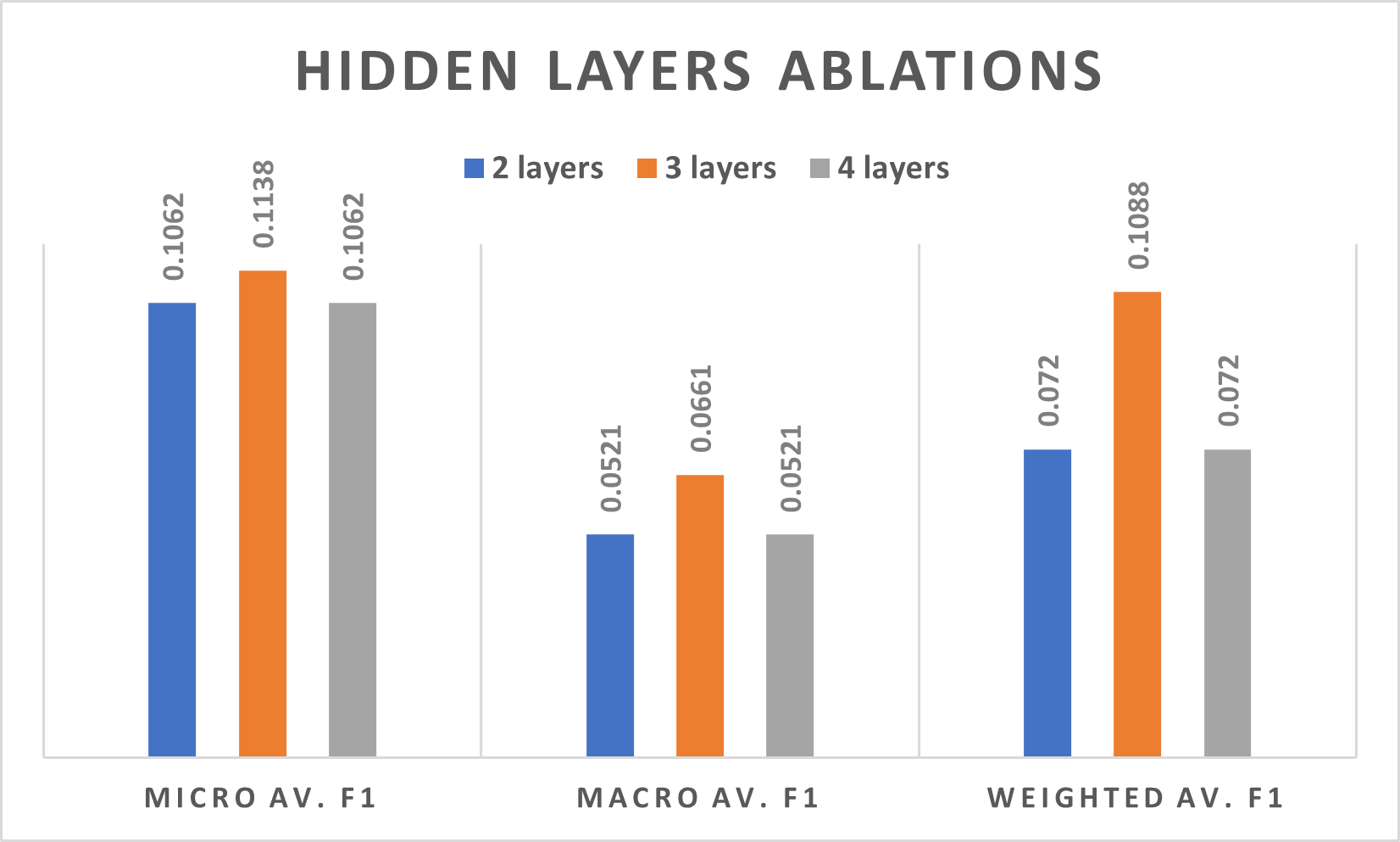}
    \caption{Ablation Study on the Birds Dataset: Effect of number of hidden layers on overall multi-label classification performance.}
    \label{fig: abs_fig}
\end{figure}

Figure \ref{fig: abs_fig} shows the effect of the number of hidden layers on overall classification performance on the Birds dataset. The best performance is obtained for three hidden layers. The model is unable to learn the underlying patterns in the data with fewer layers, whereas with more layers, it tends to overfit the data, resulting in reduced performance in these scenarios. Thus, all MLL experiments are carried out with a model having three hidden layers, except for the Flags and the FoodTruck datasets, which are relatively simpler.

$\mathcal{L}_{fmm}$ can be viewed as a product of two competing objectives - the importance factor $\exp(-(\mathbf{\zeta_k-b}))$ and $||\mathbf{\zeta_k-b}||^2$. Minimizing one leads to maximizing the other. Learning in a bipolar network requires minimizing $||\mathbf{\zeta_k-b}||^2$ \cite{EBN-MSL}. $\exp(-(\mathbf{\zeta_k-b}))$ is a stronger objective than $||\mathbf{\zeta_k-b}||^2$. Therefore, allowing gradients to flow through it during training will lead to maximizing $||\mathbf{\zeta_k-b}||^2$ to minimize the overall loss $\mathcal{L}_{fmm}$, potentially resulting in overly confident predictions and degraded test performance. Therefore, $\exp(-(\mathbf{\zeta_k-b}))$ should only be treated as a weighting factor, and gradients should not be allowed to flow through it. Table \ref{tab: abs_table} shows the effect of allowing gradients to flow through the importance factor $\exp(-(\mathbf{\zeta_k-b}))$ during training. This experiment is carried out on the Flags, Plant, Birds, Human, and FoodTruck datasets. As expected, better classification performance is obtained by stopping gradients through $\exp(-(\mathbf{\zeta_k-b}))$ during training. Hence, $\exp(-(\mathbf{\zeta_k-b}))$ is kept trainable in the remainder of the experiments.

\begin{table}[!t]
\centering
\caption{Ablation Study: Effect of stopping gradient flow through the importance factor $\exp(-(\mathbf{\zeta_k-b}))$ during training.}
\label{tab: abs_table}
\begin{tabular}{ccccc}
\hline
\multicolumn{2}{l}{\textbf{}}                    & \textbf{\begin{tabular}[c]{@{}c@{}}Micro \\ av. F1\end{tabular}} & \textbf{\begin{tabular}[c]{@{}c@{}}Macro \\ av. F1\end{tabular}} & \textbf{\begin{tabular}[c]{@{}c@{}}Weighted \\ av. F1\end{tabular}} \\ \hline
\multirow{2}{*}{\textbf{Flags}}     & with grad. & 0.4669                                                           & 0.2158                                                           & 0.294                                                               \\
                                    & w/o grad.  & \textbf{0.6063}                                                  & \textbf{0.5164}                                                  & \textbf{0.6131}                                                     \\ \hline
\multirow{2}{*}{\textbf{Plant}}    & with grad. & 0.15                                                             & 0.0924                                                           & 0.1762                                                     \\
                                    & w/o grad.  & \textbf{0.1782}                                                  & \textbf{0.1015}                                                  & \textbf{0.2066}                                                     \\ \hline
\multirow{2}{*}{\textbf{Birds}}     & with grad. & 0.0771                                                           & 0.0381                                                           & 0.0469                                                              \\
                                    & w/o grad.  & \textbf{0.1138}                                                  & \textbf{0.0661}                                                  & \textbf{0.1088}                                                     \\ \hline
\multirow{2}{*}{\textbf{Human}}     & with grad. & 0.1552                                                           & 0.1175                                                           & 0.2306                                                              \\
                                    & w/o grad.  & \textbf{0.1631}                                                  & \textbf{0.1441}                                                  & \textbf{0.2353}                                                     \\ \hline
\multirow{2}{*}{\textbf{FoodTruck}} & with grad. & 0.1355                                                           & 0.0523                                                           & 0.0373                                                              \\
                                    & w/o grad.  & \textbf{0.3401}                                                  & \textbf{0.1629}                                                  & \textbf{0.3784}                                                     \\ \hline
\end{tabular}
\end{table}

\subsection{Multi-Label Learning Performance}
Table \ref{tab: mll_perf} presents the comparison of performance on MLL for the two loss functions. In most cases, $\mathcal{L}_{fmm}$ outperforms $\mathcal{L}_{mm}$. The maximum improvement in the micro-averaged F1 score (approximately 0.23) is attained on the Yeast dataset. DOSA trained with $\mathcal{L}_{fmm}$ shows the maximum improvement on the macro, and the weighted averaged F1 scores (approximately 0.23 and 0.27, respectively) on the Flags dataset. 

\begin{table}[!t]
\centering
\caption{MLL Performance: Micro, Macro, and Weighted averaged F1 scores.}
\label{tab: mll_perf}
\begin{tabular}{ccccc}
\hline
\multicolumn{2}{c}{\textbf{}}               & \textbf{Micro}  & \textbf{Macro}  & \textbf{Weighted} \\ \hline
\multirow{2}{*}{\textbf{Yeast}}     & $\mathcal{L}_{mm}$ & 0.1783          & 0.1125          & 0.0988            \\
                                    & $\mathcal{L}_{fmm}$ & \textbf{0.4053} & \textbf{0.1937} & \textbf{0.2787}   \\ \hline
\multirow{2}{*}{\textbf{Flags}}     & $\mathcal{L}_{mm}$ & 0.4951          & 0.2856          & 0.3412            \\
                                    & $\mathcal{L}_{fmm}$ & \textbf{0.6063} & \textbf{0.5164} & \textbf{0.6131}   \\ \hline
\multirow{2}{*}{\textbf{Plant}}     & $\mathcal{L}_{mm}$ & 0.0878          & 0.0546          & 0.0495            \\
                                    & $\mathcal{L}_{fmm}$ & \textbf{0.1782} & \textbf{0.1015} & \textbf{0.2066}   \\ \hline
\multirow{2}{*}{\textbf{Birds}}     & $\mathcal{L}_{mm}$ & 0.1071          & 0.0621          & 0.1044            \\
                                    & $\mathcal{L}_{fmm}$ & \textbf{0.1138} & \textbf{0.0661} & \textbf{0.1088}   \\ \hline
\multirow{2}{*}{\textbf{Eukaryote}} & $\mathcal{L}_{mm}$ & 0.0837          & 0.0739          & 0.2096            \\
                                    & $\mathcal{L}_{fmm}$ & \textbf{0.1102} & \textbf{0.0767} & \textbf{0.2661}   \\ \hline
\multirow{2}{*}{\textbf{Human}}     & $\mathcal{L}_{mm}$ & 0.1377          & 0.1297          & 0.2088            \\
                                    & $\mathcal{L}_{fmm}$ & \textbf{0.1631} & \textbf{0.1441} & \textbf{0.2353}   \\ \hline
\multirow{2}{*}{\textbf{Gnegative}} & $\mathcal{L}_{mm}$ & 0.1302          & \textbf{0.101}  & 0.1617            \\
                                    & $\mathcal{L}_{fmm}$ & \textbf{0.3273} & 0.0941          & \textbf{0.2344}   \\ \hline
\end{tabular}
\end{table}

Table \ref{tab: imb_perf} shows the performance of DOSA on the most imbalanced class and the inverse weighted average F1 score on all classes. The most imbalanced class is the one with the least number of samples belonging to it. On the datasets shown, $\mathcal{L}_{fmm}$ improves the performance on the most imbalanced classes over $\mathcal{L}_{mm}$. These improvements add up across all (or most) imbalanced classes, considerably improving the inverse weighted average F1 score (Table 3). This underlies high F1 scores attained by $\mathcal{L}_{fmm}$ on the datasets shown in table \ref{tab: mll_perf}. The trainable margin vector $\mathbf{b}$ and the importance factor $\exp(-\mathbf{\zeta_k-b})$ in the loss function could be the reasons for the overall better performance of $\mathcal{L}_{fmm}$.

\begin{table}[!t]
\centering
\caption{Performance on the most imbalanced class and inv. weighted averaged F1 scores on all classes.}
\label{tab: imb_perf}
\begin{tabular}{ccccc}
\hline
\multirow{2}{*}{\textbf{}} & \multicolumn{2}{c}{\textbf{\begin{tabular}[c]{@{}c@{}}F1 score on\\ most imbalanced\\ class\end{tabular}}} & \multicolumn{2}{c}{\textbf{\begin{tabular}[c]{@{}c@{}}inv. Weighted av.\\ F1 score on\\ all classes\end{tabular}}} \\ \cline{2-5} 
                           & $\mathcal{L}_{mm}$                                           & $\mathcal{L}_{fmm}$                                                    & $\mathcal{L}_{mm}$                                               & $\mathcal{L}_{fmm}$                                                        \\ \hline
\textbf{Yeast}             & 0.0872                                          & \textbf{0.9128}                                          & 0.3815                                              & \textbf{0.6469}                                              \\
\textbf{Emotions}          & 0.2921                                          & \textbf{0.7079}                                          & 0.4141                                              & \textbf{0.5859}                                              \\
\textbf{Gpositive}         & 0.0337                                          & \textbf{0.9663}                                          & 0.2087                                              & \textbf{0.8923}                                              \\
\textbf{Human}             & 0.0161                                          & \textbf{0.9839}                                          & 0.0744                                              & \textbf{0.4858}                                              \\ \hline
\end{tabular}
\end{table}

\begin{figure*}[!t]
\centering
\begin{subfigure}
\centering
    \includegraphics[width=0.45\textwidth]{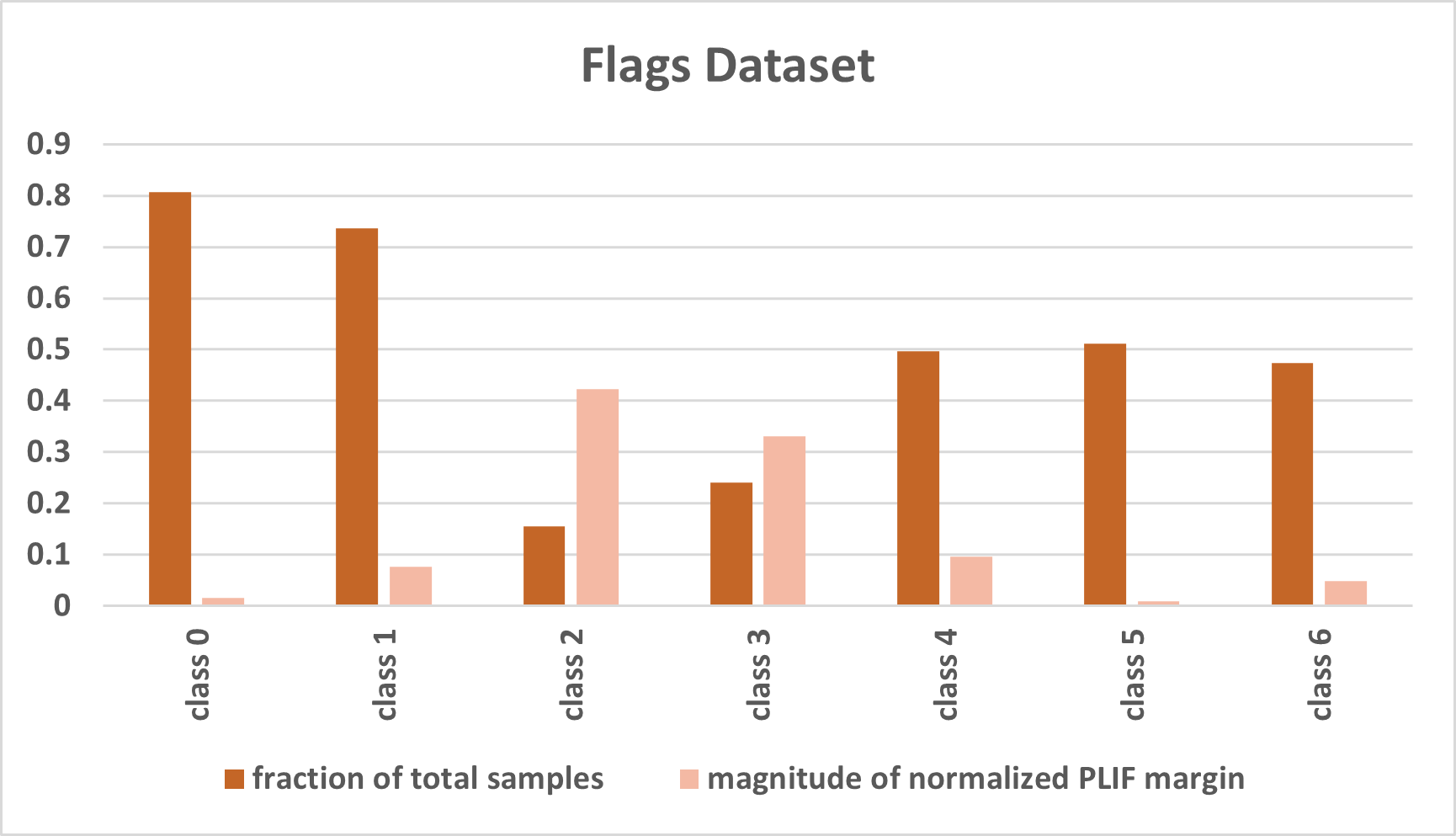}    
\end{subfigure}%
\begin{subfigure}
\centering
    \includegraphics[width=0.45\textwidth]{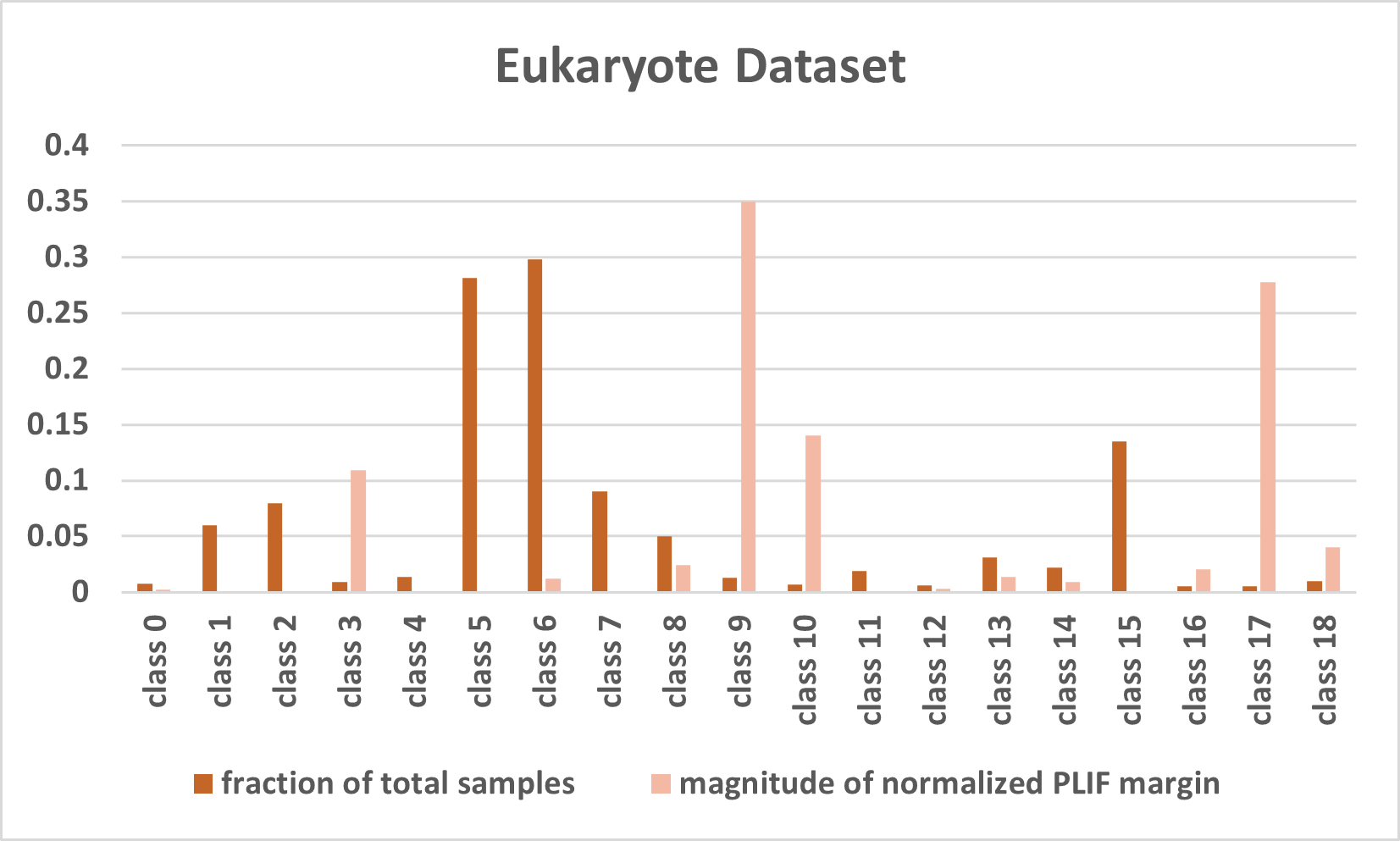}
\end{subfigure}\\
\caption{Proportion of Samples and Normalized margin values. PLIF denotes parametric leaky integrate and fire neuron (used at the hidden layers) \cite{PLIF}.}
\label{fig: corr}
\end{figure*}

Figure \ref{fig: corr} shows the proportion of samples $p_k$ and the normalized margin values $b_k$ for each class $k$ as bar plots for the Flags and the Eukaryote datasets. The normalized margin $b_k$ is defined as

\begin{equation}
    b_k = \frac{|b_k|}{\sum_j |b_j|}
\end{equation}

It is observed that classes having fewer samples (thus, larger imbalance) exhibit larger values of margin and vice versa. This shows that the learnable margin adapts itself to the imbalance present in the classes. The model confidence is initially low for these classes, resulting in a high loss value for them. As training proceeds, the model makes more confident predictions on these classes, and the corresponding margin value also increases to minimize the loss. The opposite effect takes place for classes with more samples (less imbalance), preventing the model from making overly confident predictions on them. 

\subsection{Continual Multi-Label Learning Performance}
\begin{figure*}[!t]
\centering
\begin{subfigure}
\centering
    \includegraphics[width=0.3\textwidth]{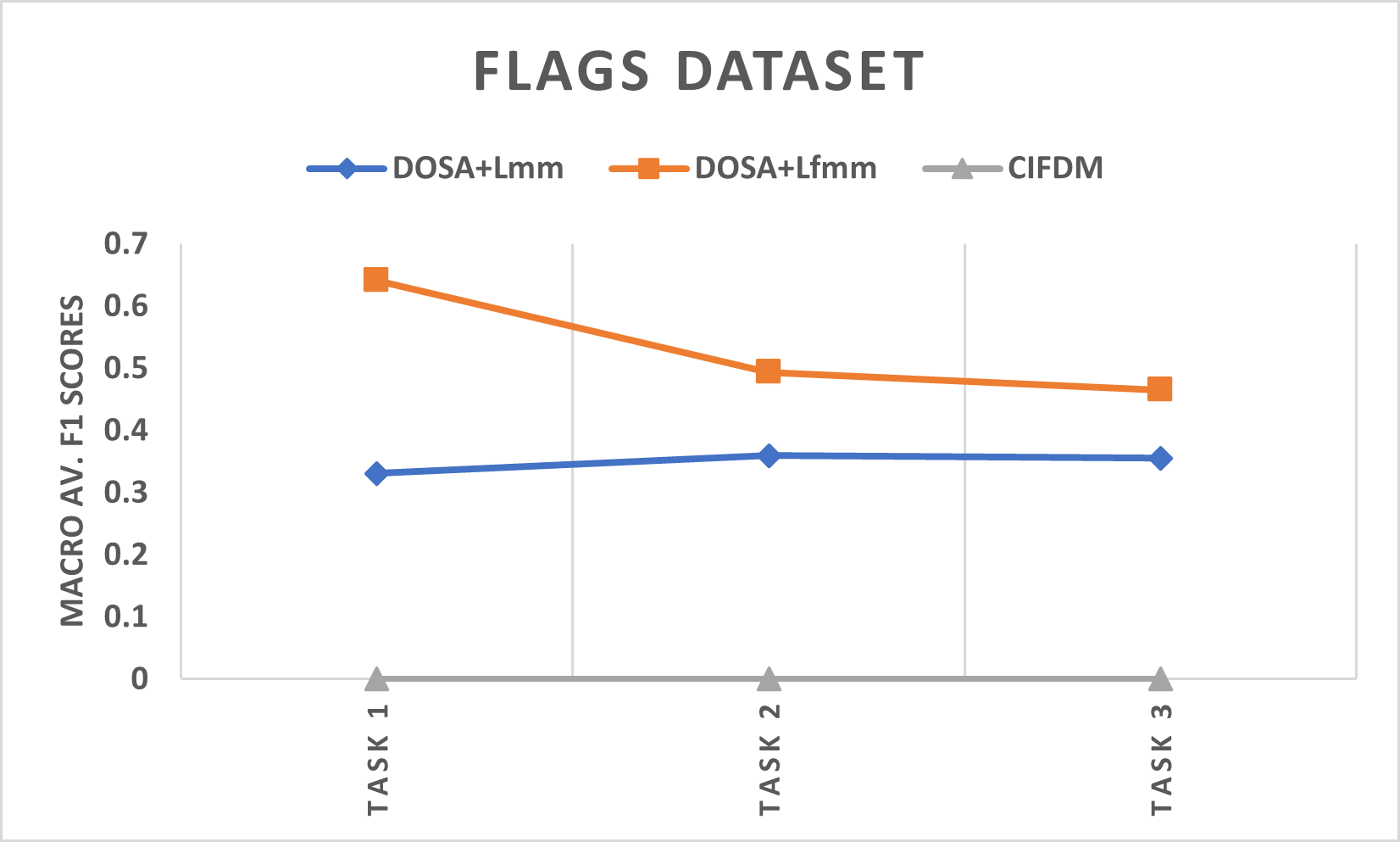}    
\end{subfigure}%
\begin{subfigure}
\centering
    \includegraphics[width=0.3\textwidth]{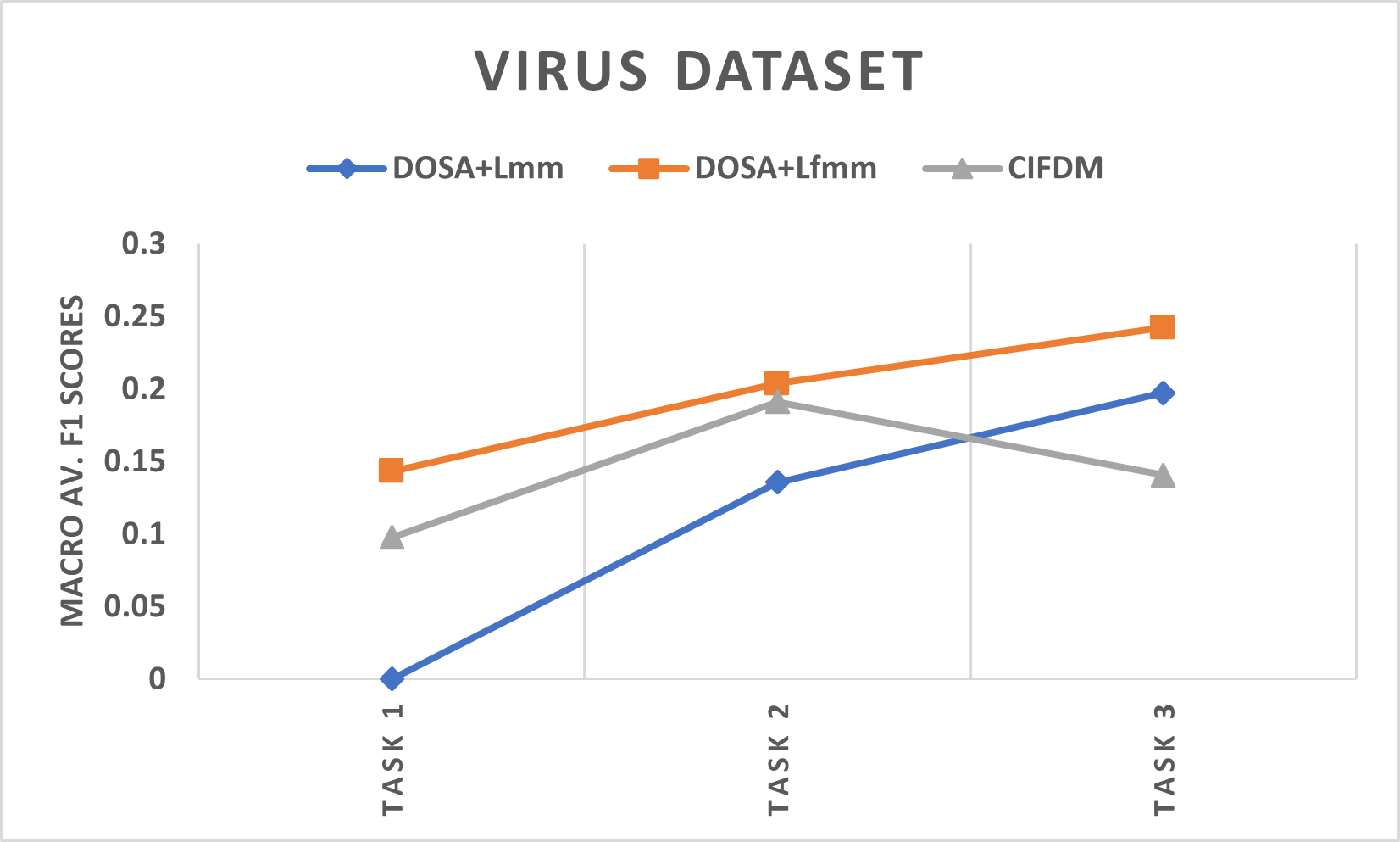}
\end{subfigure}%
\begin{subfigure}
\centering
    \includegraphics[width=0.3\textwidth]{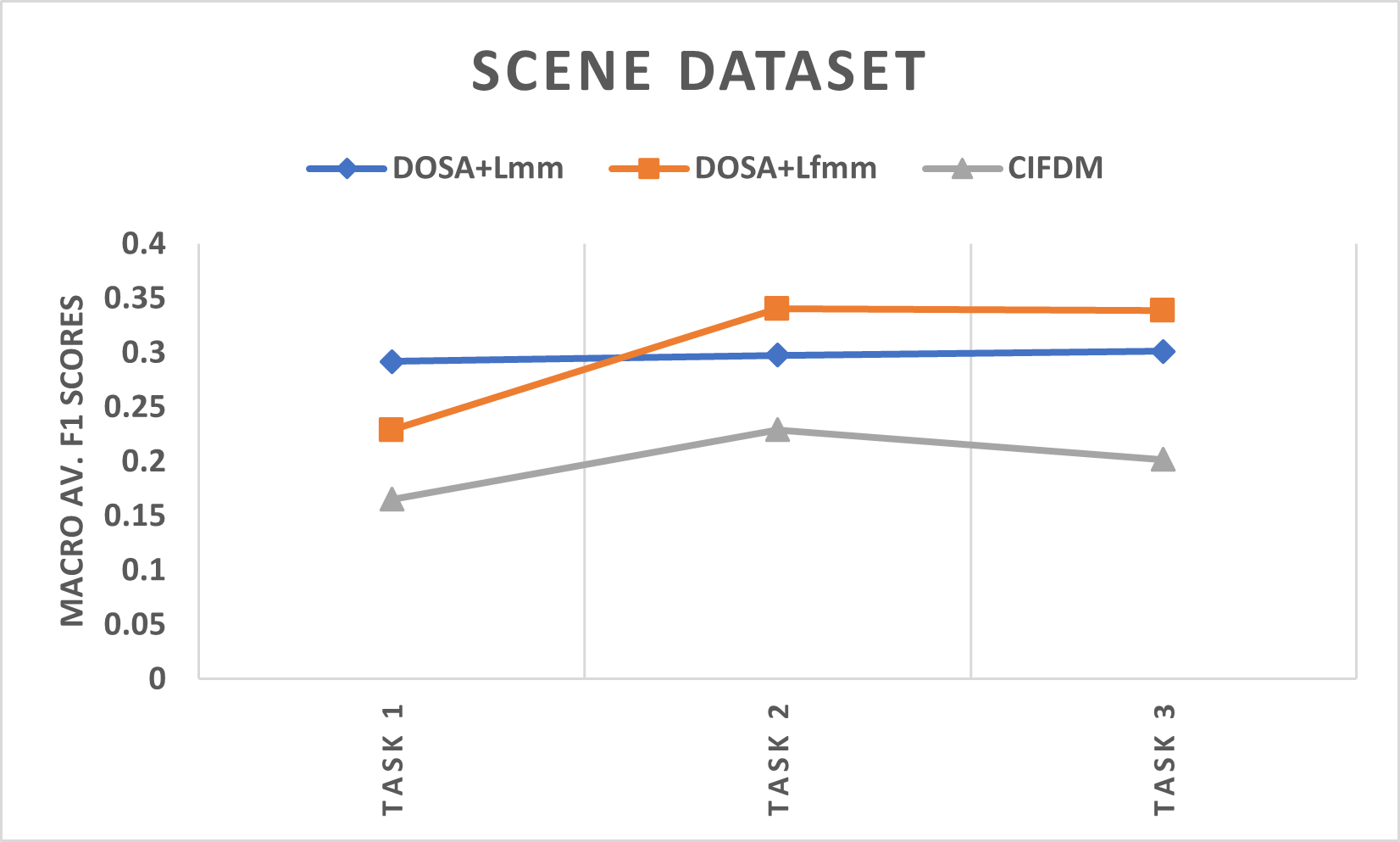}
\end{subfigure}\\
\begin{subfigure}
\centering
    \includegraphics[width=0.45\textwidth]{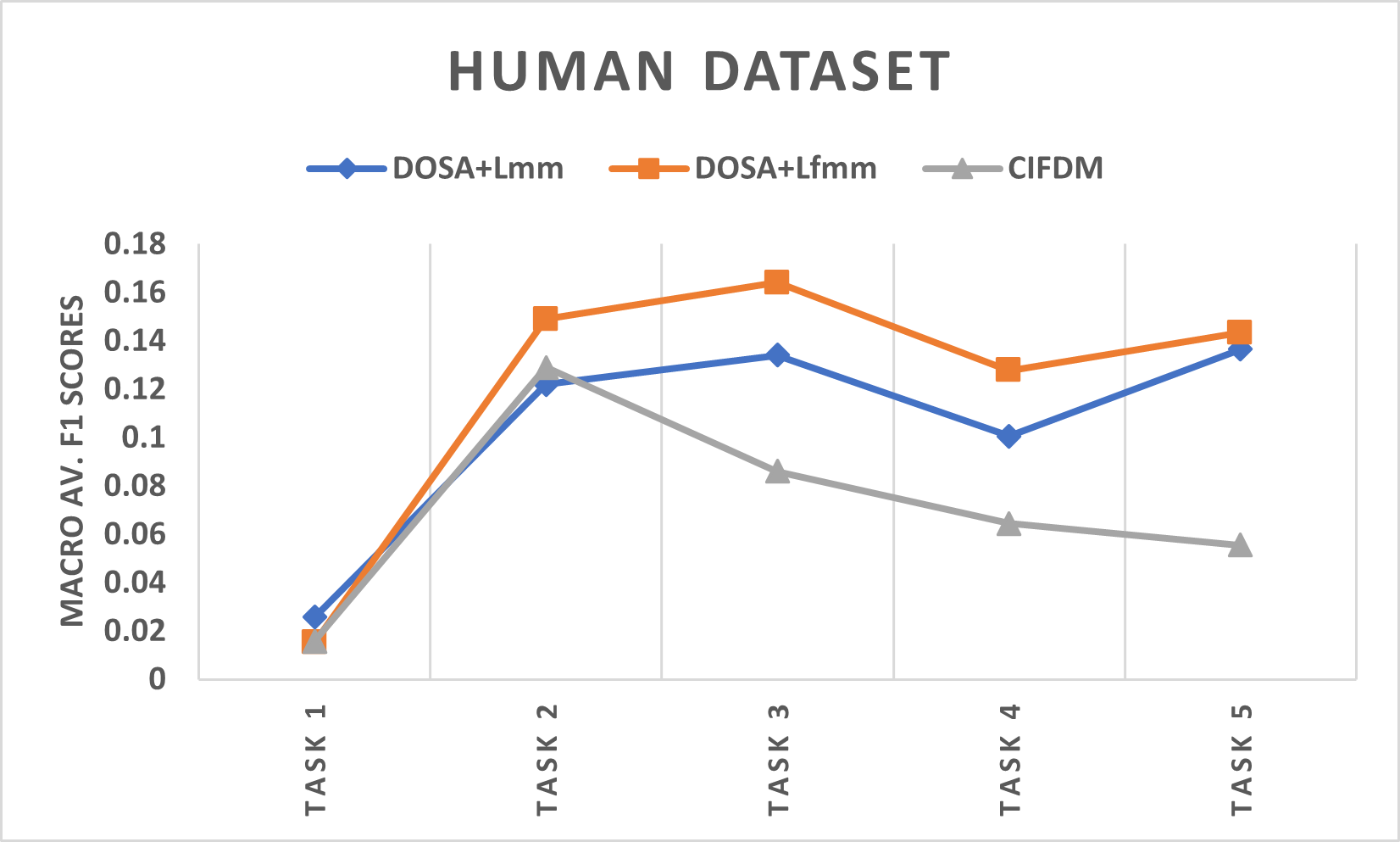}
\end{subfigure}%
\begin{subfigure}
\centering
    \includegraphics[width=0.45\textwidth]{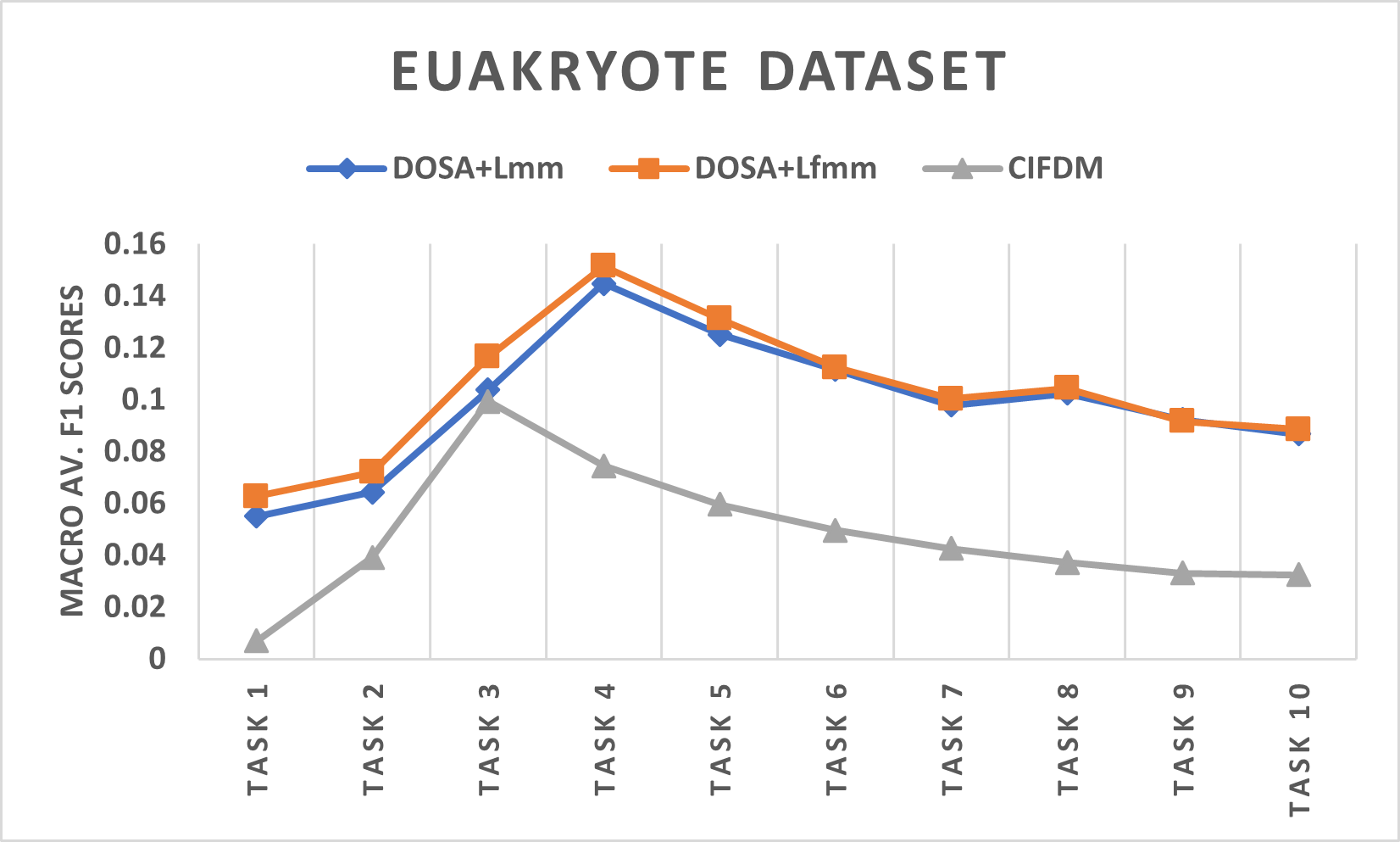}
\end{subfigure}\\
\caption{CMLL Performance Comparison: DOSA with maximum margin loss ($\mathcal{L}_{mm}$), with focal maximum margin loss ($\mathcal{L}_{fmm}$), and CIFDM.}
\label{fig: Figure 1}
\end{figure*}

Figure \ref{fig: Figure 1} shows the performance of the DOSA on CMLL. The model is trained in the SEA scheme and is evaluated in the combined mode after it is trained on all tasks. The macro-averaged F1 score is used for evaluating model performance. With $\mathcal{L}_{fmm}$, DOSA performs the best in all cases. The most significant improvements are seen for the Flags dataset across all tasks. However, CIFDM \cite{cifdm} finds it difficult to learn the categorical features present in the Flags dataset, yielding very low F1 scores. For the eukaryote dataset, $\mathcal{L}_{fmm}$ performs only slightly better than $\mathcal{L}_{mm}$. The Eukaryote dataset is relatively larger than other datasets and comprises more tasks, increasing chances for catastrophic forgetting. In the SEA mode \cite{EBN-MSL}, there is no separate mechanism (like distillation between models trained on consecutive tasks) to address catastrophic forgetting directly. This could explain the small improvements seen in the case of the eukaryote dataset. On the scene and the human datasets, $\mathcal{L}_{fmm}$ achieves better performance than $\mathcal{L}_{mm}$ on all tasks except the first one. On the virus dataset, $\mathcal{L}_{fmm}$ achieves better performance than $\mathcal{L}_{mm}$ on all tasks.

\section{Conclusions and Future Work}
This work addresses task-agnostic CMLL with SNNs by using dual output SNN architecture (DOSA) for accurate multi-label classification. Existing loss functions for bipolar networks do not incorporate the model's classification confidence measures to address data imbalance. To bridge this gap, this work proposes an imbalance-sensitive focal loss function with a trainable margin for each class ($\mathcal{L}_{fmm}$). Experiments on several benchmark datasets show the effectiveness of $\mathcal{L}_{fmm}$ in handling imbalance. $\mathcal{L}_{fmm}$ yields significant improvements over $\mathcal{L}_{mm}$ \cite{EBN-MSL}. The Yeast and the Flags datasets yield the maximum improvements on the micro and the macro averaged F1 scores, respectively. With $\mathcal{L}_{fmm}$, DOSA also achieves better CMLL performance than $\mathcal{L}_{mm}$ and CIFDM \cite{cifdm}. $\mathcal{L}_{fmm}$ performs better than $\mathcal{L}_{mm}$, possibly due to the trainable margin and the capacity to differentiate between samples based on the model's ability to classify them during learning. 

Knowledge distillation can be incorporated to enhance model performance across tasks. Learning loss function online has shown to improve model performance \cite{loss_online} on online learning tasks. It might further improve the robustness of DOSA to data imbalance. Future research will explore these directions and develop a proper framework for model training and evaluation in CMLL tasks.


\bibliographystyle{IEEEbib}
\bibliography{refs}

\end{document}